\documentclass{article}
\usepackage[utf8]{inputenc}
\usepackage{amsmath}
\usepackage{graphicx}
\usepackage{geometry}
\usepackage{url} 
\usepackage{float}
\usepackage{siunitx}
\usepackage{caption}
\usepackage{subcaption}

\title{A Parametric Bi-Directional Curvature-Based Framework for Image Artifact Classification and Quantification}
\author{Diego Frias \\ Bahia State University (UNEB), Salvador, Bahia, Brazil}
\date{\today}

\begin{document}

\maketitle

\begin{abstract}
This work presents a novel framework for No-Reference Image Quality Assessment (NR-IQA) founded on the analysis of directional image curvature. Within this framework, we define a measure of Anisotropic Texture Richness (ATR), which is computed at the pixel level using two tunable thresholds—one permissive and one restrictive—that quantify orthogonal texture suppression. When its parameters are optimized for a specific artifact, the resulting ATR score serves as a high-performance quality metric, achieving Spearman correlations ($\rho$) with human perception of approximately -0.93 for Gaussian blur and -0.95 for white noise on the LIVE dataset. The primary contribution is a two-stage system that leverages the differential response of ATR to various distortions. First, the system utilizes the signature from two specialist ATR configurations to classify the primary artifact type (blur vs. noise) with over 97\% accuracy. Second, following classification, it employs a dedicated regression model mapping the relevant ATR score to a quality rating to quantify the degradation. On a combined dataset, the complete system predicts human scores with a coefficient of determination ($R^2$) of 0.892 and a Root Mean Square Error (RMSE) of 5.17 DMOS points. This error corresponds to just 7.4\% of the dataset's total quality range, demonstrating high predictive accuracy. This establishes our framework as a robust, dual-purpose tool for the classification and subsequent quantification of image degradation.
\end{abstract}

\section{Introduction}

Automatic Image Quality Assessment (IQA) is a fundamental research field with critical applications, ranging from the optimization of compression codecs to quality control in medical imaging systems, such as the removal of ring artifacts in computed tomography sinograms. Many No-Reference IQA (NR-IQA) metrics are designed to quantify a specific class of distortion. However, practical systems often require not only to quantify the loss of quality but also to diagnose its cause.

This work addresses this gap by proposing a unified and parametric framework based on directional curvature analysis. The main contributions are:
\begin{itemize}
    \item The formulation of an "Anisotropic Textural Richness" (ATR) metric and the demonstration that it can be optimized to become a state-of-the-art predictor for the perception of both blur and white noise.
    \item The discovery that the response signature of two specialist ATR filters can be used to classify the dominant distortion type with high accuracy.
    \item The development of a complete hybrid system that first classifies and then quantifies the degradation, rigorously validated on the LIVE dataset.
\end{itemize}

\section{Related Work}
\label{sec:related_work}

Image Quality Assessment (IQA) is a well-established research field, with methods classified into three categories: Full-Reference (FR), Reduced-Reference (RR), and No-Reference (NR). FR metrics, such as SSIM \cite{Wang2004SSIM}, achieve high correlation with human perception but require access to the pristine source image, which is impractical in many applications. Consequently, research has focused on NR-IQA methods.

NR-IQA methods often rely on Natural Scene Statistics (NSS). The premise is that high-quality images possess predictable statistical properties, which distortions alter. The seminal BRISQUE method \cite{Mittal2012BRISQUE} exemplifies this approach by modeling the distribution of normalized contrast coefficients. Following this line, NIQE \cite{Mittal2013NIQE} eliminated the need for training on human scores by constructing a "quality" model from a corpus of pristine images. Recent research continues to refine NSS models; for instance, Venkataramanan and Bovik \cite{Venkataramanan2024RelaxedNSS} proposed a "relaxed" NSS model that loosens statistical assumptions to better fit a wider range of image content.

While many NSS models operate on first-order derivatives (gradients), the analysis of second-order derivatives (curvature) can capture finer textures. Sinno and Bovik \cite{Sinno2018SecondOrder} established the relevance of the second derivative for IQA by proposing a second-order NSS model that explores the correlation between curvature statistics and perceptual quality. This principle is central to our work. The utility of NSS has also been shown for specific diagnostic tasks, such as noise estimation, where Gupta et al. \cite{Gupta2018Noise} demonstrated that additive noise predictably alters the distribution of image gradients.

Advances in Deep Learning have significantly driven the NR-IQA field. The PaQ-2-PiQ method \cite{Ying2020PaQ2PiQ}, for instance, trained a CNN to assess both technical and aesthetic quality, showcasing the ability of neural networks to learn complex representations of perceptual quality. Recognizing that different artifacts require distinct models, DBCNN \cite{Zhang2020DBCNN} utilizes two parallel networks: one to predict the distortion type and another to predict the quality. Similarly, the FUNQUE framework \cite{Venkataramanan2022FUNQUE} proposes the fusion of multiple evaluators to obtain a more robust prediction, consolidating the idea that combining "specialists" is a powerful approach.

Despite the success of existing methods, a gap remains. Deep learning models often function as "black boxes," requiring vast datasets and computational power. On the other hand, classic feature-based methods, while lighter and more interpretable, typically yield a single quality score without providing diagnostic insights into the nature of the degradation. The ability to not only quantify but also classify the artifact type using a lightweight framework remains an area of interest.

This work addresses this gap by proposing a unified and parametric framework that, through second-order curvature analysis, can be optimized to create specialist filters. It is shown that the response signature of these filters can be used to build a hybrid system that first classifies the artifact (blur vs. white noise) and then quantifies its severity with a specialized regression model, offering a more complete and diagnostic quality analysis.

\section{The Parametric Curvature Framework}
\label{sec:method}
The proposed method quantifies image features through a three-stage process: curvature computation, normalization, and conditional filtering.

\subsection{Curvature Maps and Normalization}

The core of the method is the measurement of the second derivative of the image intensity. For an image $I(x, y)$, horizontal ($C_h$) and vertical ($C_v$) curvature maps are computed via convolution with 1D Laplacian kernels, $K_h = \begin{bmatrix} 1 & -2 & 1 \end{bmatrix}$ and $K_v = K_h^T$.

The resulting curvature maps exhibit a heavy-tailed distribution, where most pixels in homogeneous regions have values near zero, while a few pixels on sharp edges present very high magnitude values. To mitigate the disproportionate impact of these outliers on subsequent statistical calculations, a logarithmic transform is applied to compress the dynamic range of the values. The \texttt{log1p} function, equivalent to $\ln(1 + |C|)$, is used, yielding the normalized maps:
\begin{equation}
    L_h = \ln(1 + |C_h|) \quad \text{and} \quad L_v = \ln(1 + |C_v|)
\end{equation}

This logarithmic normalization step serves two essential functions. First, it stabilizes the variance of the data, ensuring that the standard deviation calculation is not dominated by a few extreme values, which would otherwise lead to unrealistic detection thresholds. Second, the logarithmic transformation aligns the signal processing with principles of human visual perception, as described by the Weber-Fechner Law \cite{Fechner1860}, which posits that the perception of a stimulus is approximately proportional to the logarithm of its intensity.

Finally, the standard deviations of the normalized maps, $\sigma_h = \text{std}(L_h)$ and $\sigma_v = \text{std}(L_v)$, are computed. These values serve as robust reference statistics of the image's textural dispersion, which are fundamental for the adaptive filtering stage.

\subsection{Parametric Filtering and Operating Modes}
The filtering stage classifies pixels based on their edge orientation. A horizontal edge in the image corresponds to a high rate of change in the vertical direction, while a \textbf{vertical edge} corresponds to a high rate of change in the horizontal direction. The masks for each orientation, $M_{horz}$ and $M_{vert}$, are defined by the following conditions:
\begin{align}
    M_{horz}(x, y) = 1 \iff L_v(x, y) \geq \beta \cdot \sigma_v \land L_h(x, y) < \alpha \cdot \sigma_h \\
    M_{vert}(x, y) = 1 \iff L_h(x, y) \geq \beta \cdot \sigma_h \land L_v(x, y) < \alpha \cdot \sigma_v
\end{align}
where $\alpha$ is a tolerance parameter and $\beta$ is an activation threshold.

The relationship between $\alpha$ and $\beta$ fundamentally defines the filter's behavior, creating a two-stage selection process. The $\alpha$ parameter acts as a pre-selection criterion, defining the pool of candidate pixels based on their directional purity (how low the curvature is in the orthogonal direction). The $\beta$ parameter then acts as the final selection criterion, choosing from this pool the pixels that possess sufficient strength in the principal direction. This interaction enables two distinct operating modes, whose properties are summarized in Table~\ref{tab:filter_modes}.

\begin{table}[H]
    \centering
    \caption{Comparison of the framework's two operating modes.}
    \label{tab:filter_modes}
    \renewcommand{\arraystretch}{1.2}
    \begin{tabular}{p{0.25\textwidth}|p{0.3\textwidth}|p{0.3\textwidth}}
        \hline \hline
        \textbf{Property} & \textbf{Saliency Mode ($\beta > \alpha$)} & \textbf{Texture Mode ($\alpha > \beta$)} \\
        \hline
        Dominant Parameter & $\alpha$ (Tolerance / Purity) & $\beta$ (Activation / Strength) \\
        \small{Pre-selection Criterion ($\alpha$)} & Strict and Selective & Permissive and Tolerant \\
        \small{Final Selection Criterion ($\beta$)} & Validator on a small pool & Main selector on a large pool \\
        Detected Features & Pure and strong lines (edges) & Complex and strong textures \\
        Resulting Mask & Sparse & Dense \\
        \small{Intersection ($M_{vert} \cap M_{horz}$)} & Empty & Non-empty (corners, textures) \\
        Primary Application & Structure Extraction & Quality Assessment \\
        \hline \hline
    \end{tabular}
\end{table}

In Saliency Extractor Mode, the $\beta > \alpha$ condition imposes a strict purity criterion. The filter is dominated by the $\alpha$ parameter, selecting only pixels with extreme anisotropy. The result is a sparse mask, ideal for structural analysis.

In contrast, in Texture Detector Mode, the $\alpha > \beta$ condition makes the purity criterion much more tolerant. The filter is now dominated by the $\beta$ parameter, which selects from a vast pool of complex textures those that possess sufficient strength. The result is a dense mask, highly sensitive to the overall textural richness of the image, which proved ideal for quality assessment. The following sections will empirically validate these two operating modes.

\section{Experimental Setup}
\subsection{Dataset and Performance Metrics}
\begin{figure}[H]
    \centering
    \includegraphics[width=0.95\linewidth]{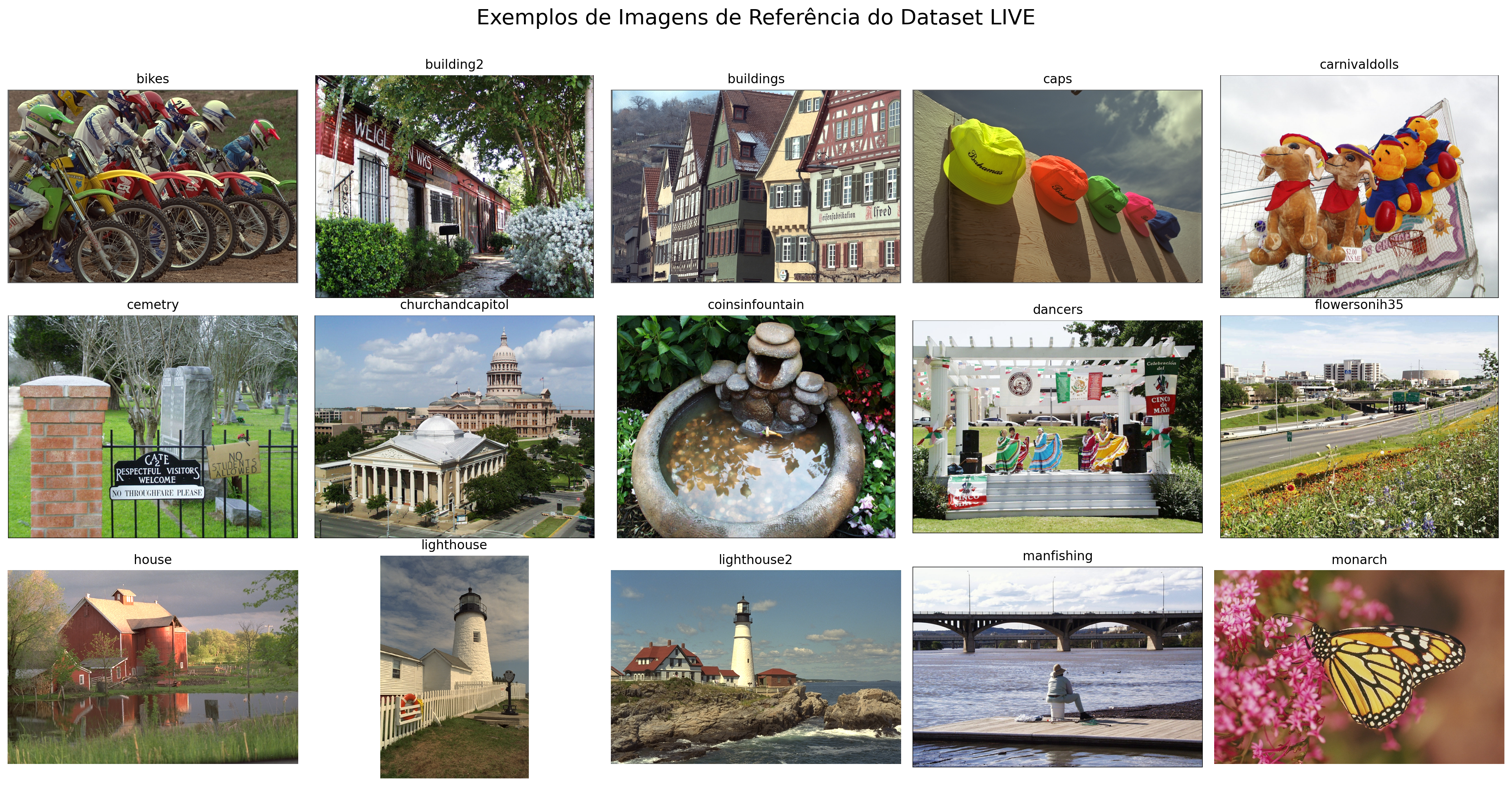}
    \caption{A sample of 15 reference images from the LIVE dataset, illustrating the diversity of content used for training and validation. The scenes include natural landscapes, man-made structures, people, and close-ups with fine textures.}
    \label{fig:live_references}
\end{figure}

All experiments were conducted on the public LIVE Image Quality Assessment Database (Release 2) \cite{Sheikh2005LIVE}. This dataset contains 29 reference images of diverse content, along with their degraded versions across five distortion types. This study focused on the "Gaussian blur" (gblur) and "White noise" (wn) subsets. Figure~\ref{fig:live_references} shows a sample of the reference images used, illustrating the variety of scenes, textures, and structures present in the dataset.

Performance was evaluated against the ground truth Differential Mean Opinion Scores (DMOS) provided with the dataset, using four standard metrics: Spearman's Rank Correlation Coefficient ($\rho$), the Coefficient of Determination ($R^2$), Root Mean Squared Error (RMSE), and Mean Absolute Error (MAE).

\section{Results}

\subsection{Optimization and Validation of Specialist Filters}\label{sec:optim_and_validation}

To specialize the framework for distinct distortions, two expert filters were derived by independently optimizing the hyperparameters $(\alpha, \beta)$ on the blur and white noise datasets. The optimization objective was to maximize the absolute Spearman's Rank Correlation Coefficient ($|\rho|$) between the ATR metric score and the ground truth DMOS values. A Grid Search was performed over a parameter space for $\alpha$ and $\beta$.

The search identified the following optimal parameter sets:
\begin{itemize}
    \item \textbf{$F_{blur}$:} Calibrated for Gaussian blur, with parameters $(\alpha=4.0, \beta=2.5)$.
    \item \textbf{$F_{noise}$:} Calibrated for white noise, with parameters $(\alpha=1.5, \beta=1.0)$.
\end{itemize}

To assess the robustness and generalization capability of these optimized filters, a K-fold cross-validation procedure ($K=5$) was carried out. Table~\ref{tab:cross_validation_results} summarizes the mean and standard deviation of the performance metrics across the test folds for each specialist filter.

\begin{table}[H]
    \centering
    \caption{Cross-validated performance (Mean $\pm$ SD) of the specialist filters ($K=5$).}
    \label{tab:cross_validation_results}
    \begin{tabular}{l S[table-format=-1.3(3)] S[table-format=-1.3(3)]}
        \hline \hline
        \textbf{Performance Metric} & {\textbf{$F_{blur}$ (vs. Blur)}} & {\textbf{$F_{noise}$ (vs. Noise)}} \\
         & {$(\alpha=4.0, \beta=2.5)$} & {$(\alpha=1.5, \beta=1.0)$} \\ \hline
        Spearman's $\rho$    & {-0.933 $\pm$ 0.037} & {-0.948 $\pm$ 0.014} \\
        Pearson's $r$        & {-0.741$\pm$ 0.044} & {-0.894 $\pm$ 0.016} \\
        RMSE (DMOS points)   & {6.414 $\pm$ 1.112}  & {4.986 $\pm$ 0.730}  \\
                             & {(10.25\%)}      & {(8.67\%)}       \\
        MAE (DMOS points)    & {4.798 $\pm$ 0.319}  & {4.009 $\pm$ 0.549}  \\
                             & {(7.67\%)}       & {(6.97\%)}       \\
        \hline \hline
    \end{tabular}
\end{table}

The analysis of the cross-validation results reveals several key findings:

\begin{itemize}
    \item \textbf{High Overall Performance:} Both filters demonstrate a very strong monotonic correlation with human perception, with mean Spearman coefficients of -0.933 for $F_{blur}$ and -0.948 for $F_{noise}$. This validates the framework's effectiveness in quantifying both types of degradation.
    
    \item \textbf{Consistency and Generalization:} The low standard deviations, particularly for the Spearman correlation (0.037 for blur and a notable 0.014 for noise), indicate that the filters' performance is highly consistent across different image contents, confirming the method's robustness and generalization capability.
    
    \item \textbf{Difference in Linearity:} The $F_{noise}$ filter exhibits a significantly stronger Pearson correlation ($r=-0.894$) than that of the $F_{blur}$ filter ($r=-0.741$). This suggests that the relationship between the $ATR_{noise}$ score and the perception of noise is more linear than the relationship between the $ATR_{blur}$ score and the perception of blur.
    
\end{itemize}

In summary, the validation confirms that it is possible to optimize high-performance specialist filters for different distortion classes from the same framework. The stability of both filters establishes the foundation for constructing a hybrid diagnostic system, as detailed in Section~\ref{sec:hybrid_system}.

\subsubsection{Analysis of Parameter Optimization Results}

The finding that the "Texture Detector" mode ($\alpha > \beta$) significantly outperformed the pure edge extraction configurations for the IQA task is a central result of this work. This provides robust quantitative evidence for a fundamental hypothesis in visual perception: that human assessment of sharpness is more closely related to the "preservation of textural richness" than to the acutance of isolated edges.

This conclusion is consistent with theories based on Natural Scene Statistics (NSS), which posit that the human visual system is optimized to process the complex, high-frequency textures abundant in natural environments \cite{Olshausen1997}. Blur degradation acts as a low-pass filter, precisely removing these textural details, which results in an image whose statistics drastically diverge from the natural norm, thus being perceived as low quality \cite{Moorthy2011BIQI}. The success of the optimized filter $(\alpha=4.0, \beta=2.5)$, which is by design tolerant to complexity (high $\alpha$) and sensitive to textural strength (moderate $\beta$), demonstrates that it acts as an effective quantifier of this "structural integrity." Therefore, the high correlation of the ATR metric with human scores is not a coincidence, but a consequence of its alignment with fundamental principles of visual perception.

\subsection{DMOS Predictive Models}\label{sec:predictive_models}

A hybrid regression model was developed which applies a predictive model trained specifically for the detected artifact class. Data analysis revealed that a second-degree polynomial model in the log-log space captures the relationship between ATR and DMOS with greater accuracy. The predictive model is therefore defined as:
\begin{equation}
\widehat{\text{DMOS}} = \exp\left(c_T + b_{1,T} \cdot \ln(ATR_T) + b_{2,T} \cdot (\ln(ATR_T))^2\right)
\end{equation}
where the subscript $T$ indicates that the coefficients $(c, b_1, b_2)$ are specialists for the artifact type $T \in \{\text{blur, noise}\}$. Table~\ref{tab:regression_models} presents the coefficients and performance of each specialist model, fitted independently on its respective dataset.

\begin{figure}[H]
    \centering
    \includegraphics[width=0.49\linewidth]{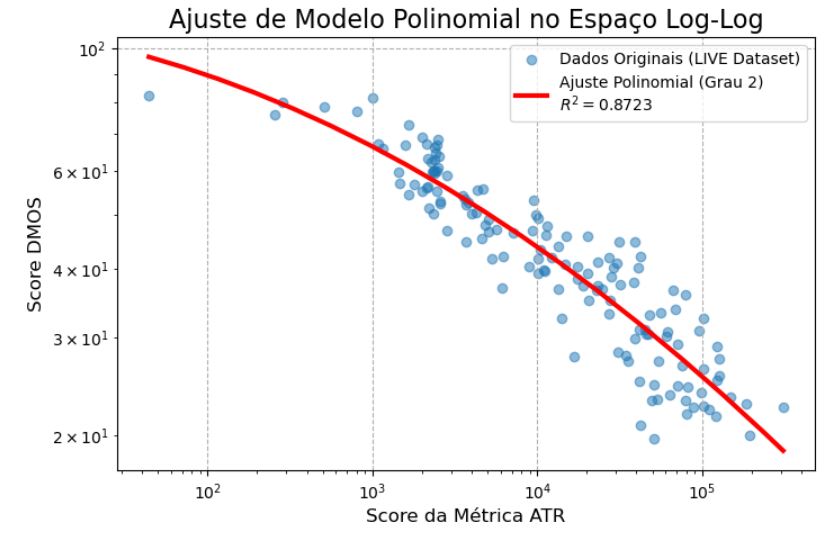}
    \hfill 
    \includegraphics[width=0.49\linewidth]{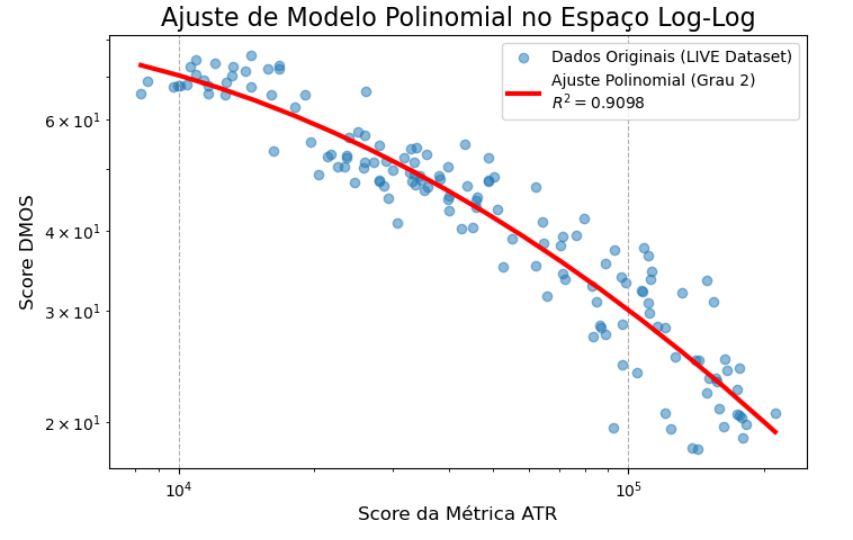}
    \caption{Scatter plots of experimental data (ATR vs. DMOS) in log-log scale. \textbf{Left:} Specialist model fit for the \textbf{blur} dataset ($R^2=0.872$). \textbf{Right:} Specialist model fit for the \textbf{white noise} dataset ($R^2=0.910$). The red lines represent the fitted second-degree polynomial curves.}
    \label{fig:model_plots}
\end{figure}

Figure~\ref{fig:model_plots} visually illustrates the strong adherence of both models' curves to their respective experimental data, confirming the suitability of the polynomial approach.

\begin{table}[H]
    \centering
    \caption{Coefficients and performance of the specialist regression models (2nd-degree polynomial). Percentage errors are relative to the DMOS dynamic range of each subset (62.58 for blur, 57.51 for noise).}
    \label{tab:regression_models}
    \sisetup{
        round-mode=places,
        round-precision=4,
        table-format=-1.4
    }
    \renewcommand{\arraystretch}{1.3}
    \begin{tabular}{l|ccccc}
        \hline \hline
        \textbf{Specialist Model} & 
        {\textbf{Coef. $c$}} & 
        {\textbf{Coef. $b_1$}} & 
        {\textbf{Coef. $b_2$}} & 
        {\textbf{$R^2$}} & 
        {\textbf{RMSE (\%)}} \\ \hline
        Blur (`vs. gblur`) & 4.7232 & 0.0027 & -0.0114 & 0.8723 & \num{5.5052} (8.80\%) \\
        Noise (`vs. wn`)   & 0.0526 & 1.1162 & -0.0717 & 0.9098 & \num{4.7946} (8.34\%) \\
        \hline \hline
    \end{tabular}
\end{table}

Both models achieved a high degree of fit and prediction power. The specialist model for noise explained 91.0\% of the data variance ($R^2=0.9098$). The prediction errors, when normalized by the DMOS dynamic range in each dataset, are consistently low. The RMSE for the blur model (5.51 points) represents 8.80\% of the score range, while the RMSE for the noise model (4.79 points) corresponds to 8.34\% of its respective range. These low relative errors confirm the accuracy and practical utility of both specialist models.

\subsection{Hybrid System for Artifact Classification and Quantification}\label{sec:hybrid_system}

With the two specialist filters validated, the possibility of using them in conjunction to create a hybrid system capable of first classifying the artifact type and then quantifying its magnitude was investigated.

\subsubsection{Artifact Classification in Feature Space}

It was observed that the response of the two filters forms a unique signature. For images with blur, $ATR_{noise} > ATR_{blur}$, whereas for images with white noise, $ATR_{blur} > ATR_{noise}$. 

\begin{figure}[H]
    \centering
    \includegraphics[width=0.65\textwidth]{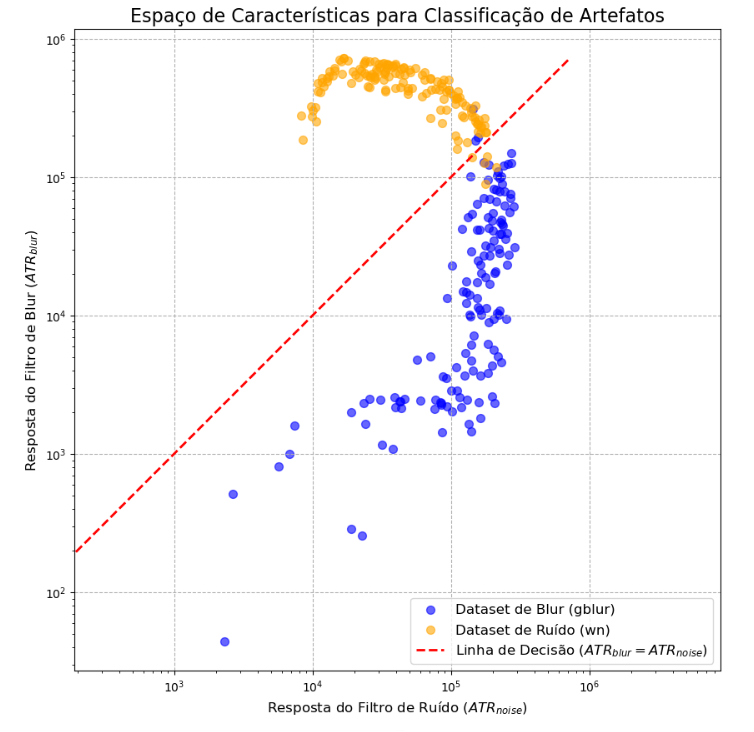}
    \caption{Feature space of the scores. Points from the gblur (blue) and wn (orange) datasets occupy distinct regions. The identity line $y=x$ (dashed) acts as an effective decision boundary.}
    \label{fig:classifier}
\end{figure}

This simple decision rule functions as an artifact classifier. To visualize the diagnostic capability, each distorted image from the blur and noise subsets was mapped as a point in a 2D feature space, with axes defined by the $ATR_{noise}$ and $ATR_{blur}$ scores. The result, presented in Figure~\ref{fig:classifier}, demonstrates a near-perfect separability between the two classes.

The analysis of the trajectories in Figure~\ref{fig:classifier} reveals a distinct dynamic for each degradation. Starting from a common high-response region for the reference images, the trajectory of \textbf{blur} (blue) is characterized by a monotonic decrease, causing the points to move into the region where $ATR_{noise} > ATR_{blur}$. In contrast, the trajectory of \textbf{white noise} (orange) exhibits an initial response where $ATR_{blur}$ increases, moving the points into the region where $ATR_{blur} > ATR_{noise}$. This unique response signature allows for the formulation of a simple decision rule with a classification accuracy exceeding 97\%:
\begin{equation}
\text{Artifact Type} = \begin{cases} 
      \text{Blur} & \text{if } ATR_{noise} > ATR_{blur} \\
      \text{Noise} & \text{if } ATR_{blur} \geq ATR_{noise}
   \end{cases}
\end{equation}

\subsubsection{End-to-End Predictive System Performance}

Based on this classification rule, a hybrid regression model was implemented that first identifies the dominant noise type and then applies the corresponding specialist model\footnote{Described in Section~\ref{sec:predictive_models}} to predict the DMOS. 

\begin{figure}[H]
    \centering
    \includegraphics[width=0.7\linewidth]{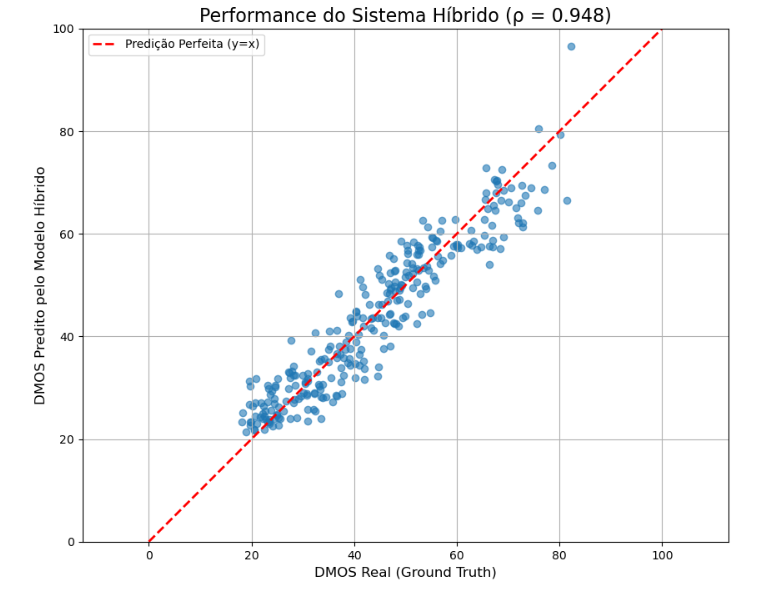}
    \caption{Scatter plot of the final hybrid system's performance. Each point represents an image from the combined dataset (blur and noise). The horizontal axis shows the ground truth DMOS (human perception), and the vertical axis shows the model's predicted DMOS. The strong concentration of points around the identity line ($y=x$, dashed) visually illustrates the high correlation (Spearman's $\rho=-0.948$) and low prediction error of the system.}
    \label{fig:end_to_end_fit}
\end{figure}

The complete system was evaluated on the combined dataset (blur and noise). Table~\ref{tab:final_results} presents the final end-to-end performance of the system, and Figure~\ref{fig:end_to_end_fit} visually illustrates this high performance, showing the strong concentration of the predicted points around the identity line.

\begin{table}[H]
    \centering
    \caption{Final performance of the hybrid system, evaluated on the combined dataset. Percentage errors in parentheses are relative to the total DMOS dynamic range (64.12 points).}
    \label{tab:final_results}
    \sisetup{round-mode=places, round-precision=4}
    \renewcommand{\arraystretch}{1.3}
    \begin{tabular}{lc}
        \hline \hline
        \textbf{Performance Metric} & \textbf{Value} \\ \hline
        Pearson's $r$ & \num{-0.9442} \\
        Spearman's $\rho$ & \num{-0.9478} \\
        Coefficient of Determination ($R^2$) & \num{0.8916} \\
        RMSE (DMOS points) & \num{5.1729} (8.07\%) \\
        MAE (DMOS points) & \num{4.1505} (6.47\%) \\
        \hline \hline
    \end{tabular}
\end{table}

The results confirm the high performance of the system. The Spearman coefficient of -0.948 and the low prediction error (RMSE of 5.17 DMOS points) validate the effectiveness of the hybrid approach, which combines artifact classification with specialized quantification.

\section{Discussion}
The framework's performance can be attributed to its adaptive thresholding mechanism. Blur degradation reduces the ATR score by removing textural complexity, while noise degradation reduces it by introducing chaotic variance that inflates the detection thresholds. The system effectively learned to measure "structural integrity," a fundamental attribute of image quality. The ability to diagnose the artifact via the response signature of the specialist filters is the main contribution, as it enables the application of more accurate regression models.

\begin{figure}[H]
    \centering
    \includegraphics[width=0.75\linewidth]{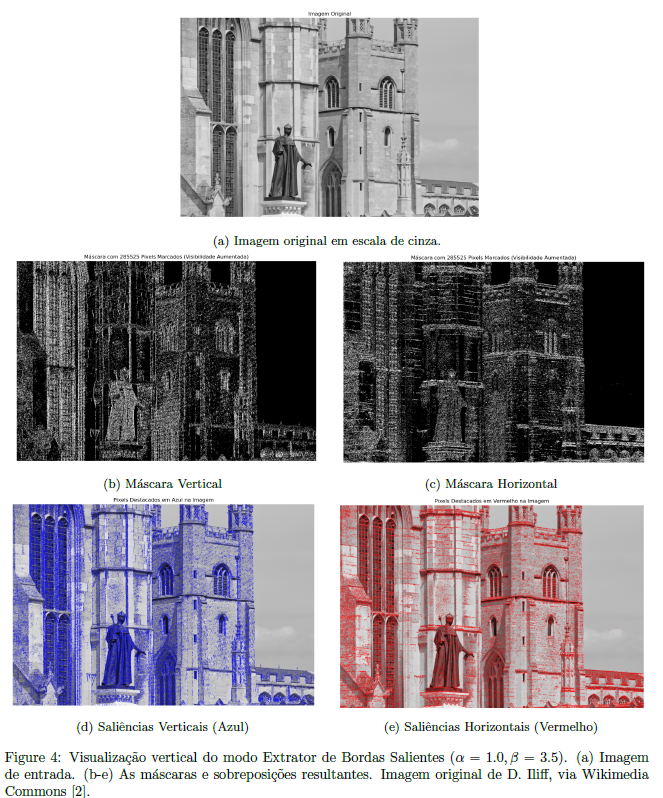}
    \caption{Visualization of the Saliency Extractor mode ($\alpha=1.0, \beta=3.5$). (a) Input image. (b-e) The resulting masks and overlays. Original image by D. Iliff, via Wikimedia Commons \cite{Iliff2013KingsCollege}.}
    \label{fig:saliency_mosaic}
\end{figure}

As a secondary application, the Saliency Extractor mode ($\beta \gg \alpha$) was qualitatively tested. Figure~\ref{fig:saliency_mosaic} demonstrates the filter's ability to sparsely isolate features in the primary focal plane of an architectural image \cite{Iliff2013KingsCollege}, functioning as a high-contrast saliency detector. A quantitative validation of this functionality remains as promising future work.

\section{Conclusion}
This work introduced a dual-purpose framework for image analysis. It was demonstrated that the method can be optimized to create specialist filters for different distortion types, such as blur and white noise, both achieving state-of-the-art performance ($\rho \approx -0.94$). The main contribution was the development of a hybrid system that utilizes the response signature of these filters to first classify the artifact type and then quantify its magnitude with high accuracy ($R^2=0.892$). The flexibility and validated performance of the method establish it as a robust contribution to both quality assessment and content analysis in digital images.

\bibliographystyle{plain}
\bibliography{referencias}

\end{document}